\documentclass[10pt,journal,compsoc]{IEEEtran}
\usepackage{ragged2e}
\usepackage{amsmath}
\usepackage{subfigure}
\usepackage{graphicx}
\usepackage{multirow}
\usepackage{amsthm}
\usepackage[font=bf]{caption}
\usepackage{amsfonts}
\usepackage{amssymb}

\ifCLASSOPTIONcompsoc
  \usepackage[nocompress]{cite}
\else
  \usepackage{cite}
\fi

\ifCLASSINFOpdf
\else
\fi

\begin{document}
%
\title{Deep Conversational Recommender in Travel}

\author{Lizi Liao, Ryuichi Takanobu, Yunshan Ma, Xun Yang, Minlie Huang
        and~Tat-Seng~Chua
\IEEEcompsocitemizethanks{
\IEEEcompsocthanksitem L. Liao is the corresponding author. E-mail: liaolizi.llz@gmail.com
\IEEEcompsocthanksitem R. Takanobu and M. Huang are with Tsinghua University
\IEEEcompsocthanksitem L. Liao, Y. Ma, Y. Xun and TS. Chua are with National University of Singapore.}
\thanks{Manuscript received June 13, 2019; revised **** ****.}}

%

\markboth{Journal of \LaTeX\ Class Files,~Vol.~14, No.~8, June~2019}%
{Shell \MakeLowercase{\textit{et al.}}: Bare Demo of IEEEtran.cls for Computer Society Journals}
%



\IEEEtitleabstractindextext{%
\justify
\begin{abstract}
When traveling to a foreign country, we are often in dire need of an intelligent conversational agent to provide instant and informative responses to our various queries.
However, to build such a travel agent is non-trivial. 
First of all, travel naturally involves several sub-tasks such as hotel reservation, restaurant recommendation and taxi booking \textit{etc}, which invokes the need for global topic control.
Secondly, the agent should consider various constraints like price or distance given by the user to recommend an appropriate venue. 
In this paper, we present a Deep Conversational Recommender (\textbf{DCR}) and apply to travel. It augments the sequence-to-sequence (seq2seq) models with a neural latent topic component to better guide response generation and make the training easier. To consider the various constraints for venue recommendation, we leverage a graph convolutional network (GCN) based approach to capture the relationships between different venues and the match between venue and dialog context. For response generation, we combine the topic-based component with the idea of pointer networks, which allows us to effectively incorporate recommendation results.
We perform extensive evaluation on a multi-turn task-oriented dialog dataset in travel domain and the results show that our method achieves superior performance as compared to a wide range of baselines. 
\end{abstract}

}

\maketitle

\IEEEdisplaynontitleabstractindextext

%
\IEEEpeerreviewmaketitle

\section{Introduction}
Conversational agents and travel go hand in hand. In fact, artificial intelligence is set to be a game-changer for this industry, through helping travellers and companies simplifying travel arrangements and streamlining business procedures. Currently, large companies in travel industry such as Expedia.com, KLM and Booking.com, race to launch their online chatbots. For example, in the first month after KLM lauching their bot service, their volume of Facebook messages jumped 40 percent. It is a fact that conversational agents are revolutionizing customer care, from answering questions to solving customer issues to venue recommendations \textit{etc}.

\begin{figure}[!htp]
	\centering
	\includegraphics[scale=0.48]{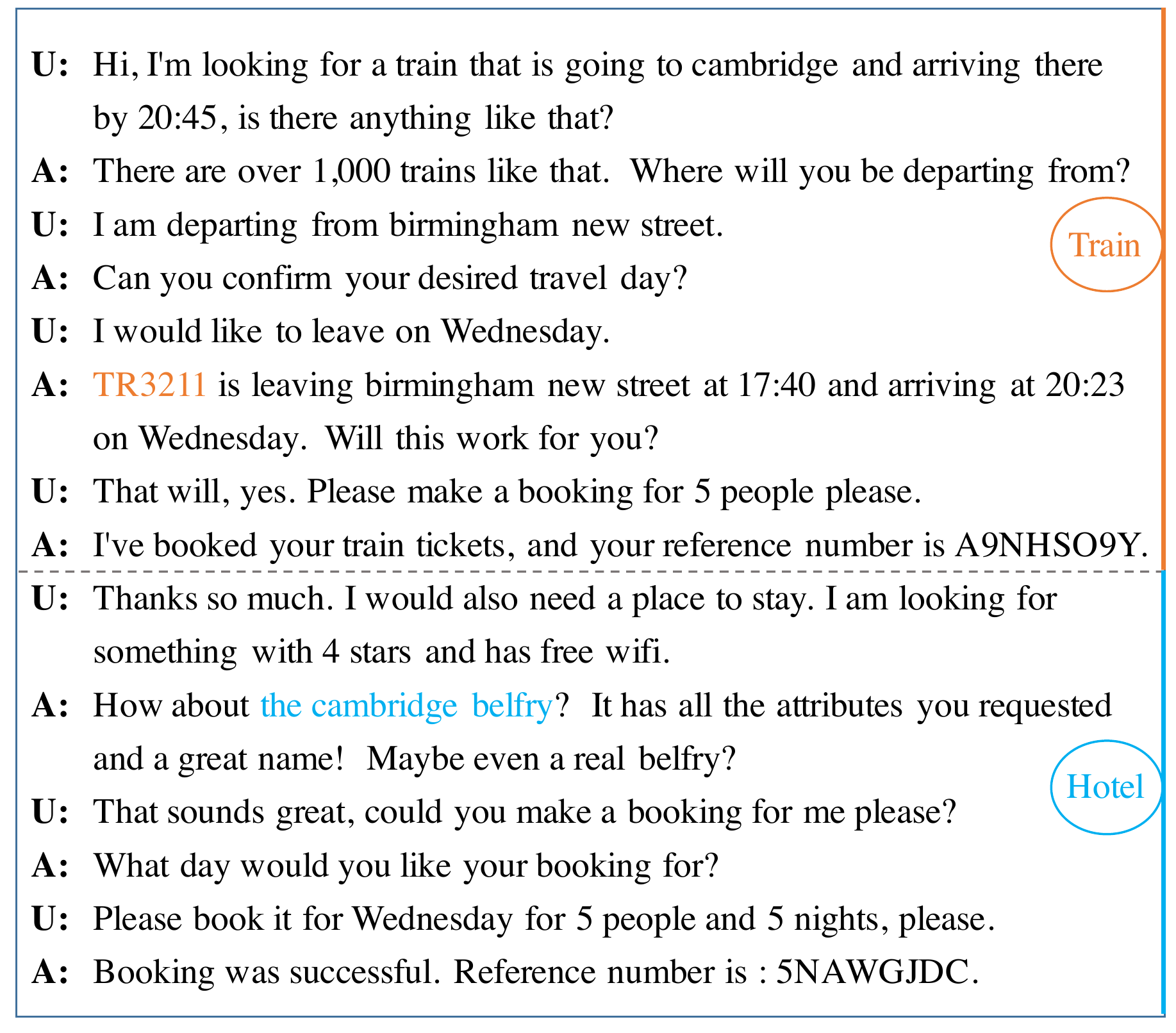}
	\caption{A sample dialog between a user (\textbf{U}) and an agent (\textbf{A}) from the dataset. We observe the need for global topic control and appropriate venue recommendation. }
	\vspace{-0.4cm}
	\label{Fig:illustration}
\end{figure}

Although conversational agents in travel show big commercial potential, it is non-trivial to build such an intelligent system to meet the various user needs. As the example illustrated in Figure \ref{Fig:illustration}, travel naturally involves several sub-tasks such as hotel reservation, restaurant booking and attraction recommendation \textit{etc}. Thus the agent should have the ability to recognize those topics from the context and generate within-topic responses. Currently, neural conversational models \cite{vinyals2015neural,shang2015neural,serban2016building} are the latest development in conversational modeling, where seq2seq-based models, such as HRED \cite{serban2016building}, are employed for generating responses in an end-to-end fashion. Such models are good at capturing the local structure of word sequence but might face difficulty in remembering global semantic structure of dialog sessions. Therefore, current state-of-the-art methods might not be sufficient for the travel agent scenario.

At the same time, to satisfy users' need such as finding hotel or restaurant, the ability to recognize constraints and efficiently leverage them in venue recommendation is essential for building an intelligent system. For example, in order to generate the hotel `cambridge belfry' as in Figure \ref{Fig:illustration}, the agent needs to capture not only the constraints like `4 stars' and `free wifi' under the same topic, but also the hidden location constraint `cambridge'. Moreover, we observe that distance and time are also important factors to consider when doing venue recommendation under the travel scenario. Indeed, there have been some existing efforts aiming to find appropriate venues in conversational systems. Many task-oriented dialog systems try to form queries and feed them to database systems to retrieve venues \cite{BordesW16,wen2017network,budzianowski2018multiwoz}. However, such methods heavily rely on the exact match of constraints which is rather sensitive to even slight language variations. It also has other limitations such as the weakness on modeling relationships between venues and the inability to back-propagate error signals from the end output to the raw inputs. To alleviate such problems, memory networks are leveraged to `softly' incorporate venue entries in external Knowledge Bases (KBs) \cite{sukhbaatar2015end,madotto2018mem2seq}. However, the various relationships between venues, even the simplest `nearby' relation, are hard to model. The most recent studies such as \cite{li2018towards,Sun2018} integrate conversational system with recommendation components, but the recommendation part only focus on learning the interplay between users and items.

\begin{figure*}[!htp]
	\centering
	\includegraphics[scale=0.66]{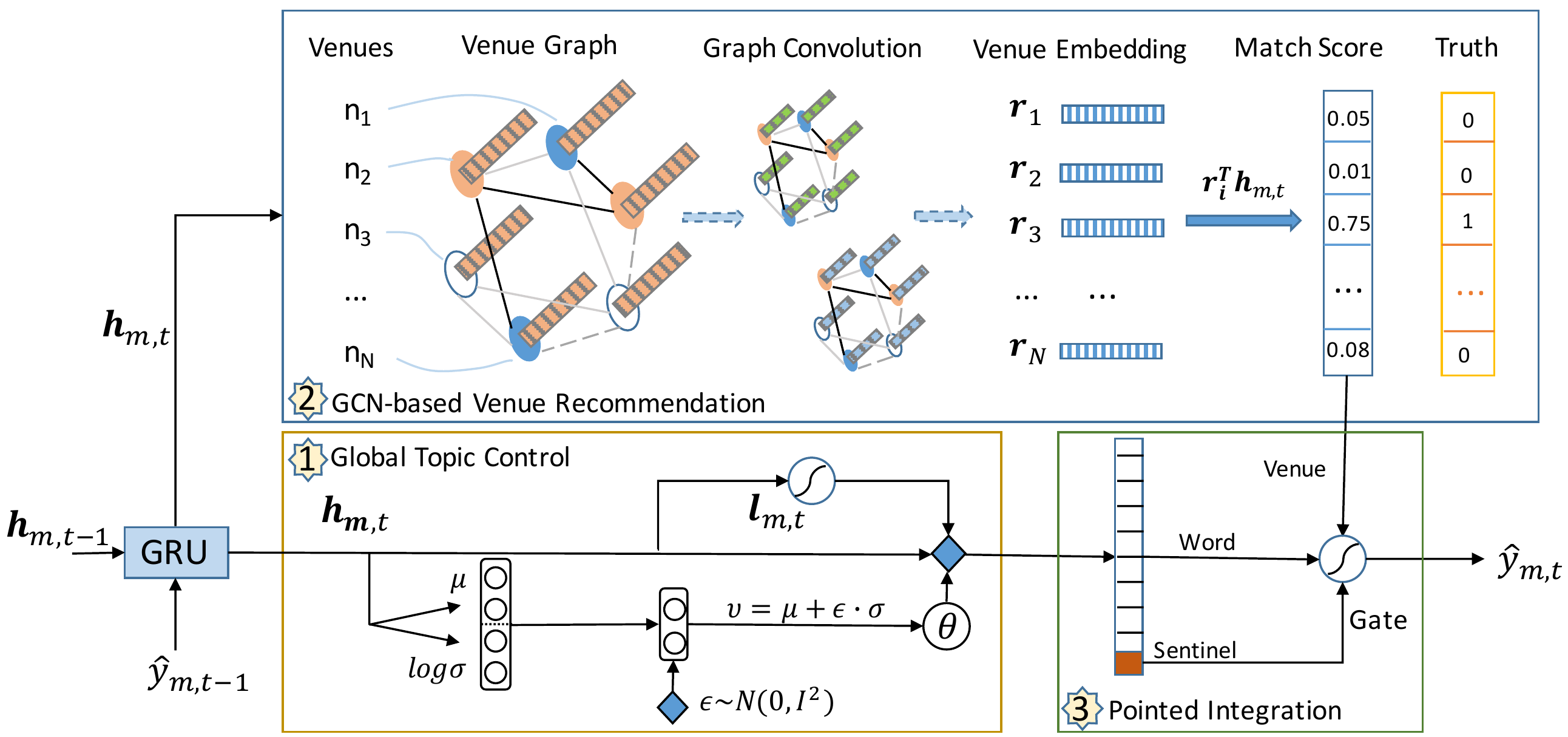}
	\caption{The proposed DCR model for travel, which consists of three components. The global topic control component enables the system to switch between various sub-tasks quickly. The GCN-based venue recommendation component generates venues by considering venue information, relations among them and the match to dialog context. Finally, a pointed integration mechanism incorporates the two components for the final response generation. The diamonds are stochastic nodes.}
	\label{Fig:model}
\end{figure*}

In this paper, we propose a Deep Conversational Recommender (\textbf{DCR}) as shown in Figure \ref{Fig:model} and apply it to the travel domain to address the above mentioned problems. First, in order to enable the agent to swiftly differentiate sub-topics in travel, we leverage the underlying seq2seq-based model to capture the local dynamics of utterances while extract and represent its global semantics by a mixture of topic components like topic models \cite{blei2003latent}.
Second, we employ a graph convolutional network (GCN) based approach to capture the various relationships between venues and learn the match between venue and dialog context. When generating venue recommendations, the agent 
ranks the venues by calculating the matching scores between the learned venue representations and dialog context representations. 
The key idea is that GCN-based component helps the conversational recommender to generate better representations of venues that incorporate both venue feature information as well as venue relations.
Third, we combine the topic-based component and the GCN-based component by leveraging the idea of pointer networks. It allows us to effectively incorporate the recommendation results into the response generation procedure.

To sum up, the main contributions of this work are threefold as follows:
\begin{itemize}
	\setlength{\itemsep}{3pt}
	\item[$\bullet$] We propose a conversational travel agent which handles multiple sub-tasks involving seven topics --- attraction, hospital, police, hotel, restaurant, taxi and train. A neural topic component helps it to generate within-topic responses by narrowing down the generation of tokens in decoding.
	\item[$\bullet$] We employ a GCN-based venue recommender which jointly captures venue information, relationships between them and the dialog contexts. Inspired by pointer networks, an integration mechanism is used to incorporate the recommendation results to the final responses.
	\item[$\bullet$] We conduct extensive experiments to evaluate the proposed method under various evaluation metrics and show superior performance over the state-of-the-art methods.
\end{itemize}

In the rest of the paper, we review related work in Section 2. Section 3 describes the elementary building blocks of the proposed learning method, including the neural latent topic component, graph convolutional network based recommender and the response integration mechanism inspired by pointer network. Experimental results and analysis are reported in Section 4, followed by conclusions and discussion of future work in Section 5.

\section{Related Work}
\subsection{Task Oriented Conversational Systems}
Task-oriented systems aim to assist users to achieve specific goals with natural language such as restaurant reservation and schedule arrangement. Traditionally, they have been built in pipelined fashion: language understanding, dialog management, knowledge query and response generation \cite{williams2007partially,hori2009statistical,young2013pomdp}. However, the requirement of human labor in designing dialog ontology and heavy reliance on slot filling as well as dialog state tracking techniques limits its usage to relatively simple and specific tasks such as flight reservation \cite{seneff2000dialogue} or querying bus information \cite{raux2005let}. For travel which involves multiple sub-tasks and needs to handle various constraints for venues recommendation, such pipelined methods might not be sufficient.

Recently, end-to-end approaches for dialog modeling, which use seq2seq-based models, have shown promising results \cite{serban2016building,wen2017network,zhao2017generative}. They directly map plain text dialog history to the output responses. Since the dialog states are latent, there is no need for hand-crafted state labels. In order to make such models generate within-topic responses, a possible way is to provide relevant database query results as a proxy for language grounding. As shown in \cite{wen2017latent}, a stochastic neural dialog model can generate diverse yet rational responses
mainly because they are heavily driven by the knowledge the model is conditioned on. However, despite the need for explicit knowledge representations, building a corresponding knowledge base and actually making use of it have been proven difficult \cite{matuszek2006introduction,miller2016key}. Therefore, progress has been made in conditioning the seq2seq model on coarse-grained knowledge representations, such as a fuzzily-matched retrieval result via attention \cite{ghazvininejad2017knowledge} or a set of pre-organized topic or scenario labels \cite{wang2017steering,xing2016topic}. In our work, we opt for a new direction to employ a hybrid of a seq2seq conversational model and a neural topic model to jointly learn the useful latent representations. Based on the learned topics, the system manages to narrow down the response generation.

\subsection{Conversational Recommender}
By offering a natural way for product or service seeking, conversational recommendation systems are attracting increasing attention. Due to the big commercial potential, companies like Amazon, Google, eBay, Alibaba are all rolling out such kind of conversational recommenders. Intuitively, integrating recommendation techniques into conversational systems can benefit both recommender and conversational systems, especially for travel. For conversational systems, good venue recommendations based on users' utterances, venue information and relations can better fulfill user's information need thus creating more business opportunities. For recommender systems, conversational systems can provide more information about user intentions, such as user preferred type of food or the location of a hotel, by interactively soliciting and identifying user intentions based on multi-round natural language conversation. 

Although conversational recommendation has shown great potential, research in this area is still at its infancy. 
Existing approaches usually are goal-oriented and combine various modules each designed and trained independently \cite{thompson2004personalized,greco2017converse}. These approaches either rely heavily on tracking the dialog state which consists of slot-value pairs, or focus on different objectives such as minimizing the number of user
queries to obtain good recommendation results. For example, \cite{li2018towards} employed user-based autoencoder for collaborative filtering and pre-trained it with MovieLens data to do recommendation. However, their recommendations are only conditioned on the movies mentioned in the same dialog, while ignores other dialog contents expressed in natural language. As another example, \cite{christakopoulou2016towards} leveraged a generative Gaussian model to recommend items to users in a conversation. However, their dialog system only asks questions about whether a user likes an item or whether the user prefers an item to another, while a typical task oriented dialog system often directly solicits facets from users \cite{dhingra2017towards,wen2017network}. 
There are also another line of approaches using reinforcement learning (RL) to train goal-oriented dialog systems \cite{li2017end,Sun2018}. For instance, in \cite{Sun2018}, a simulated user is used to help train a dialog agent to extract the facet values needed to make an appropriate recommendation. In contrast, we propose to employ a GCN-based venue recommender to take care of various constraints for venues which are prevalent in travel and seamlessly integrate these results to the response generation.

\section{The DCR Model}
We aim at building an agent capable of answering users' queries and making venue recommendations to satisfy their requirements. One might therefore characterize our system as a conversational recommender. The complete architecture of our approach is illustrated in Figure \ref{Fig:model}. Starting from the bottom of Figure \ref{Fig:model}, there are mainly three sub-components as follows.

(1) In order to help the system generate within-topic response $y_m$, a global topic control component takes in dialog context $\{u_1,\cdots,u_{m-1}\}$ with $m-1$ utterances and produces probability distribution $\textbf{p}(y_{mt})$ over each token $y_{mt}$ that favor certain topics,
\begin{align*}
\textbf{p}(y_{mt}) = f_{Topic}(\{u_1,\cdots,u_{m-1}\}|\boldsymbol{\Psi}),
\end{align*}
where $f_{Topic}$ denotes the global topic control model network and $\boldsymbol{\Psi}$ denotes the network parameters.

(2) A graph convolutional neural network based venue recommendation component learns venue representation $\textbf{R}$ by capturing various venue information and relationships. It learns the matching between dialog contexts $\{u_1,\cdots,u_{m-1}\}$ and the representations $\textbf{R}$ to generate recommendation scores $\textbf{p}$ for venues.
\begin{align*}
\textbf{p} = softmax(\textbf{R}^T\textbf{h}),
\end{align*}
where $\textbf{h}$ is the hidden representation of dialog context.

(3) The recommender's output $\textbf{p}$ is used in response generation together with the topic part output $\textbf{p}(y_{mt})$ via a pointed integration mechanism. The hard gate sentinel \$ is leveraged for choosing them.
After formalizing the problem as above, we provide more details for each of these components one by one.

\subsection{Global Topic Control}

\subsubsection{Basic Encoder}
Formally, we consider a dialog as a sequence of $M$ utterances $D=\{u_1,\cdots,u_M\}$. Each utterance $u_m$ is a sequence with $N_m$ tokens, \textit{i.e.} $u_m=\{y_{m,1},\cdots,y_{m,N_m}\}$. The $y_{m,n}$ are either tokens from a vocabulary $V$ or venue names from a set of venues $V'$. In general, seq2seq-based conversational models like \cite{serban2016building} generate a target utterance given a source utterance and dialog history. Given the dialog context $\{u_1,\cdots,u_{m-1}\}$, the goal is to produce a \textit{machine} response $u_m$ that maximizes the conditional probability $u_m^\ast=argmax_{_{u_m}} p(u_m|u_{m-1},\cdots,u_1 )$. Here, we apply the well-accepted hierarchical recurrent encoder decoder (HRED) model \cite{serban2016building} as the backbone network. At the token level, an encoder RNN maps each utterance $u_m$ to an utterance vector representation $\textbf{u}_m$, which is the hidden state obtained after the last token of the utterance has been processed. At the utterance level, a context RNN keeps track of past utterances by iteratively
processing each utterance vector and generates the hidden state $\textbf{h}_m$,
\begin{align}
p(\textbf{u}_m|\textbf{u}_{m-1},\cdots, \textbf{u}_1) &\triangleq p(\textbf{u}_m|\textbf{h}_m)\\
\textbf{h}_m &= f_{\textbf{W}_U}(\textbf{h}_{m-1},\textbf{u}_{m-1}).
\end{align}

At the token level, when the decoder of the HRED model generates tokens in \textit{machine} response $u_m$, we initialize $\textbf{h}_{m,0}= \textbf{h}_{m-1}$.
\begin{align}
p(\textbf{y}_{m,t}|\textbf{y}_{m,1:t-1}, \textbf{h}_{m-1}) &\triangleq p(\textbf{y}_{m,t}|\textbf{h}_{m,t})\\
\textbf{h}_{m,t} &= f_{\textbf{W}_H}(\textbf{h}_{m,t-1},\textbf{y}_{m,t-1})
\end{align}
where $\textbf{h}_{m,t}$ is the token level hidden state at step $t$ inside turn $m$, $f_{\textbf{W}_U}$ and $f_{\textbf{W}_H}$ are the hidden state updates that can either be a vanilla RNN cell or complex cell such as LSTM or GRU.

\subsubsection{Generative Process}
\label{generative}
While RNN-based models can theoretically model arbitrarily long dialog histories if provided enough capacity, in practice even the improved version like LSTM or GRU struggles to do so \cite{bengio1994learning,dieng2016topicrnn}. In dialogs between user and travel agent,  there usually exist long-range dependencies captured by sub-topics such as hotel reservation, restaurant finding and train ticket booking etc. Since much of the long-range dependency in language comes from semantic coherence \cite{dieng2016topicrnn}, not from syntactic structure which is more of a local phenomenon, the inability to memorize long-term dependencies prevents RNN-based models from generating within-topic responses. On the other hand, topic models are a family of models that can be used to capture global semantic coherency \cite{blei2003latent}. It relies on counting word co-occurrence to group words into groups.
Therefore, we leverage a neural topic component to extract and map between the input and output global semantics so that the seq2seq submodule can focus on perfecting local dynamics of the utterances such as the syntax and word
order.

The generative process of the global topic control component can be described as the following,
\begin{itemize}
	\setlength{\itemsep}{3pt}
	\item[1.] Encode the user input $u_{m-1}$ and dialog context $C$ into a vector representation $\textbf{h}_{m-1}=HRED(u_{m-1},\cdots,u_1) \in \mathbb{R}^d$.
	\item[2.] Draw a topic proportion vector $\boldsymbol{\theta} \thicksim N(0, \textbf{I})$.
	\item[3.] In turn $m$, initialize the decoder hidden state $\textbf{h}_{m,0} = \textbf{h}_{m-1}$.
	\item[4.] Given token $y_{m,1:t-1}$, for the $t$-th token $y_{m,t}$,	
\end{itemize}

\begin{itemize}
	\item[~]
	\begin{itemize}
		\setlength{\itemsep}{3pt}
		\item[(a)] Update the hidden state $\textbf{h}_{m,t} = f_{\textbf{W}_H}(\textbf{h}_{m,t-1},\textbf{y}_{m,t-1})$.
		\item[(b)] Draw stop word indicator $l_t \thicksim Bernoulli(sigmoid(\textbf{W}^T\textbf{h}_{m,t}))$.
		\item[(c)] Draw a token $y_{m,t} \thicksim p(y_{m,t}|\textbf{h}_{m,t},\boldsymbol{\theta},l_t,\textbf{B})$, where \\$p(y_{m,t} = i|\textbf{h}_{m,t},\boldsymbol{\theta},l_t,\textbf{B}) \varpropto exp(\textbf{w}_i^T\textbf{h}_{m,t} + (1-l_t)\textbf{b}_i^T\boldsymbol{\theta})$.
	\end{itemize}
\end{itemize}
The $HRED(\cdot)$ is the HRED model \cite{serban2016building} which encodes dialog history into a vector representation, and $N(\mu(\textbf{h}_{m-1}), \sigma^2(\textbf{h}_{m-1}))$ is a parametric isotropic Gaussian with a mean and variance both obtained from Multilayer Perceptron with input $\textbf{h}_{m-1}$ separately. The $\textbf{w}_i$ and $\textbf{b}_i$ are the corresponding columns in weight matrix $\textbf{W}$ and $\textbf{B}$. To combine with the seq2seq-based model, we adopt the hard-decision style from TopicRNN \cite{dieng2016topicrnn} by introducing a random variable $l_t$. The stop word indicator $l_t$ controls how the topic vector $\boldsymbol{\theta}$ affects the output. Note that the topic vector is used as a bias which enables us to have a clear separation of global semantics and those of local dynamics. For example, when $l_t = 1$ which indicates that $y_{m,t}$ is a stop word, the topic vector $\boldsymbol{\theta}$ will have no contribution to the output. This design is especially useful as topic models do not model stop words well, because stop words usually do not carry semantic meaning while appear frequently in almost every dialog session. 

\subsubsection{Inference}
During model inference, the observations are token sequences $u_m$ and stop word indicators $l_{1:N_m}$. The log marginal likelihood of $u_m$ is
\begin{equation}
\begin{split}
&log~ p(u_m,l_{1:N_m} | u_{1:m-1} ) = \\
&log \int_{\boldsymbol{\theta}} p(\boldsymbol{\theta}|u_{1:m-1}) \prod_{t=1}^{N_m} p(y_{m,t}|\textbf{h}_{m,t}, l_t,\boldsymbol{\theta} ) p(l_t|\textbf{h}_{m,t} ) d\boldsymbol{\theta}.
\end{split}
\label{likelihood}
\end{equation}
Since direct optimization of Equation \ref{likelihood} is intractable due to the integral over the continuous latent space, we use variational inference for approximating it \cite{jordan1999introduction}. Suppose $q(\boldsymbol{\theta}|u_{1:m})$ be the variational distribution on the marginalized variable $\boldsymbol{\theta}$, the variational lower bound of Equation \ref{likelihood} can therefore be constructed as
\begin{equation}
\label{topicloss}
\begin{split}
&\mathcal{L}(u_m,l_{1:N_m} | q(\boldsymbol{\theta}|u_{1:m}),\boldsymbol{\Psi})\\ &\triangleq \mathbb{E}_{q(\boldsymbol{\theta}|u_{1:m})} \Big[ \sum_{t=1}^{N_m} logp(y_{m,t}|\textbf{h}_{m,t}, l_t,\boldsymbol{\theta}) \\ 
&+ \sum_{t=1}^{N_m}logp(l_t|\textbf{h}_{m,t})\Big] - D_{KL}(q(\boldsymbol{\theta}|u_{1:m})||p(\boldsymbol{\theta}|u_{1:m-1})) \\
&\leq log~ p(u_m,l_{1:N_m} | u_{1:m-1},\boldsymbol{\Psi} ).
\end{split}
\end{equation}
Inspired by the neural variational inference framework in \cite{mnih2014neural,miao2016neural} and the Gaussian reparameterization trick in \cite{kingma2013auto}, we construct $q(\boldsymbol{\theta}|u_{1:m})$ as an inference network using a feed-forward neural network,
\begin{align}
q(\boldsymbol{\theta}|u_{1:m}) = N(\boldsymbol{\theta}; \mu(u_{1:m}), diag(\sigma^2(u_{1:m}))).
\end{align}
Denoting $\boldsymbol{\tau} \in \mathcal{N}_+^{|V/V_s|}$ as the term-frequency vector of $u_{1:m}$ excluding stop words (with $V_s$ as the stop word vocabulary), we have $\mu(u_{1:m}) = ReLU(\textbf{W}^T_\mu \boldsymbol{\tau})$ and $\sigma(u_{1:m}) = ReLU(\textbf{W}^T_\sigma \boldsymbol{\tau})$ where bias is omitted. Note that although $q(\boldsymbol{\theta}|u_{1:m})$ and $p(\boldsymbol{\theta}|u_{1:m-1})$ are both parameterized as Gaussian distributions, the former one only works during training while the later one generates the required topic distribution vector $\boldsymbol{\theta}$ for composing the machine response. 

Suppose during training, the one-hot vector for any token $y$ and its stop word indicator are $\textbf{y}$ and $\textbf{l}$ respectively. The predicted correspondence vectors are $\textbf{y}'$ and $\textbf{l}'$. Inspired by Equation \ref{topicloss}, the loss for this global topic control component consists of two cross entropy losses and a KL divergence between the assumed distribution and learned distribution as follows.
\begin{align}
\label{topic}
\begin{split}
\mathcal{L}_{Topic} =& ~avg.~\Big[\mathcal{L}_{cross}(\textbf{y},\textbf{y}') + \mathcal{L}_{cross}(\textbf{l},\textbf{l}')\Big]\\
& - D_{KL}(N(0, \textbf{I})||q(\boldsymbol{\theta}|u_{1:m})),
\end{split}
\end{align}
where $avg.$ indicates the averaged cross entropy loss over all training tokens.
\subsection{GCN-based Venue Recommendation}
Given the dialog context and ground truth venue node pairs, our task in this subsection is to find a good match between them. We need to leverage both the venue attributes such as `free wifi' for hotel and the various relationships between these venues. For example, when user books a hotel, he or she might also want to find a `nearby' restaurant. To jointly consider such attributes as well as the relationships, we naturally resort to graph based methods. Recently, the graph convolutional neural network (GCN) based methods have set a new standard on countless recommender system benchmarks \cite{Ying2018,hamilton2017representation}. Unlike purely content-based deep models (e.g., recurrent neural networks), GCNs leverage both content information as well as graph structure. We thus adopt the graph convolution operation into our venue recommender.

We formulate an un-directed graph structure as $G=(O,E)$, where $O=\{n_1,n_2,\cdots,n_N\}$ is a set of $N$ nodes and $E\subseteq N\times N$ is a set of edges between nodes. Here the nodes can be hotels, restaurants, location area etc, while the relations can be `nearby' or co-appear \textit{etc}. In this way, venues located in the same area will be connected closely, and venues co-appeared in the same dialog session will be connected closely. We use $\textbf{A}\in \mathbb{R}^{N\times N}$ to denote the adjacency matrix, $\widetilde{\textbf{A}} = \textbf{A} + \textbf{I}$ to denote the adjacency matrix with added self-connections and the new degree matrix $\widetilde{\textbf{D}}_{ii} = \sum_j \widetilde{\textbf{A}}_{ij}$. We denote the attributes of nodes as a matrix $\textbf{X}$, the representations of nodes in $l^{th}$ layer as $\textbf{R}^{(l)}$. Initially, we have $\textbf{R}^{(0)}=\textbf{X}$ which means that the initial representation of nodes are obtained from embedding node attributes.

\begin{figure}[!htp]
	\centering
	\includegraphics[scale=0.38]{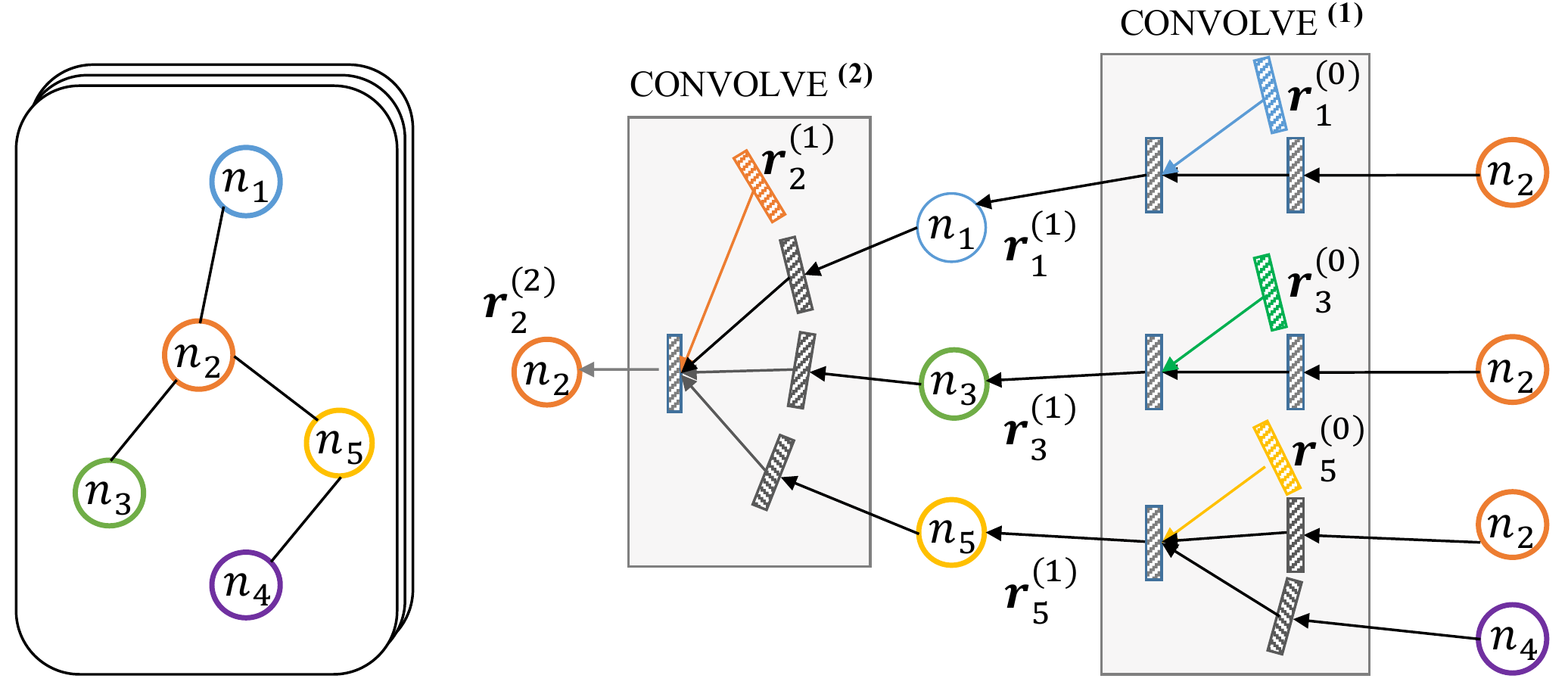}
	\caption{The illustration of convolution operation in the constructed graph. Two layers are stacked. Each $\textbf{r}^{(l)}$ denotes a node representation, corresponds to the column in $\textbf{R}^l$.}
	\label{Fig:gcn}
\end{figure}

Given such a constructed graph, we generate high-quality embeddings or representations of entities that can be used for calculating the matching score with dialog context thus obtaining the venue recommendation results. Generally speaking, to generate the embedding for a venue, we apply multiple convolutional modules that aggregate feature information from the venue's local graph neighborhood. The core idea is to learn how to iteratively aggregate feature information from local graph neighborhoods. As shown in Figure \ref{Fig:gcn},
we first project the former layer node representation $\textbf{R}^{(l-1)}$ into a latent space using the weight matrix $\textbf{W}^{(l)}$ (we omitted the bias term for simplicity),
\begin{align*}
\textbf{R}' = \textbf{R}^{(l-1)} \textbf{W}^{(l)}.
\end{align*}
Then the latent representation $\textbf{R}'$ is propagated via the normalized adjacency matrix $\widetilde{\textbf{D}}^{-\frac{1}{2}} \widetilde{\textbf{A}}\widetilde{\textbf{D}}^{-\frac{1}{2}}$ with self-connections. As demonstrated in \cite{kipf2016semi}, this propagation rule is motivated via a first-order approximation of localized spectral filters on graphs. Finally, we use the ReLU function to increase the non-linearity. Thus, a single ``convolution'' operation transforms and aggregates feature information from a node's one-hop graph neighborhood as follows,
\begin{align}
\textbf{R}^l = ReLU(\widetilde{\textbf{D}}^{-\frac{1}{2}} \widetilde{\textbf{A}}\widetilde{\textbf{D}}^{-\frac{1}{2}} \textbf{R}^{(l-1)} \textbf{W}^{(l)}).
\label{updatingrule}
\end{align}
By stacking multiple such convolutions, information can be propagated across far reaches of a graph. Here we stack two layers. 

After introducing the updating rules for node representations as in Equation \ref{updatingrule}, we present the objective function which encourages the matching between dialog context and venues. Suppose there are $M$ dialog context and ground truth node pairs, we obtain the dialog context representation $\textbf{h}_i$ and the ground truth node vector $ \textbf{s}_i \in \mathbb{R}^N$ for each pair. The objective function resumes the cross-entropy loss as follows
\begin{align}
\label{gcn}
\mathcal{L}_{GCN} =	-\frac{1}{M}\sum_{i=1}^M [\textbf{s}_i log(\textbf{p}_i) + (1-\textbf{s}_i)log(1-\textbf{p}_i)],
\end{align}
where $\textbf{p}_i = softmax(\textbf{R}^T\textbf{h}_i)$ is a vector of scores predicted by the GCN-based model, and $\textbf{R}$ is the finial node representation matrix obtained via the graph convolution process.

\subsection{Pointed Integration Mechanism}
Now, given the dialog context, we can predict the next utterance via the global topic control component and obtain the recommended venue through the GCN-based recommender. To integrate the two lines of results, we propose a pointed integration mechanism. Generally speaking, we use a Gated Recurrent Unit (GRU) \cite{chung2014empirical} to decode the system response. At each decoding step $t$ in turn $m$, the GRU gets the previously generated token and the previous hidden state as input, and generates the new hidden state,
\begin{align*}
\textbf{h}_{m,t} = GRU(\textbf{h}_{m,t-1}, \hat{\textbf{y}}_{m,t-1}).
\end{align*}
Then the hidden state $\textbf{h}_{m,t}$ is passed to two branches as shown in Figure \ref{Fig:model}. In one branch, the $\textbf{h}_{m,t}$ is passed to the global topic control component. Following the generative process introduced in Subsection \ref{generative}, the probability of generating the next token is calculated as:
\begin{align}
\textbf{p}_1(\hat{y}_{m,t})\varpropto exp(\textbf{W}^T\textbf{h}_{m,t} + (1-l_t)\textbf{B}^T\boldsymbol{\theta}).
\label{eq1}
\end{align}
In the other branch, the $\textbf{h}_{m,t}$ is fed to the GCN-based recommender. It helps the recommender rank the venues and output the top ranked venue name.
\begin{align}
\textbf{p}_2(\hat{y}_{m,t}) = softmax(\textbf{R}^T\textbf{h}_{m,t}).
\label{eq2}
\end{align}

\subsubsection{Sentinel}
In the final response generation, whether a token is generated from Equation \ref{eq1} or Equation \ref{eq2} is decided via a sentinel. As detailed before, we have a set of venue names $V'$. At the very beginning, we substitute all the venue names in dataset with the sentinel token $\$$. Thus the vocabulary for topic control component is $V$ which consists of all the tokens appearing in our dataset (expect the venue names) plus the $\$$ token. During the response decoding process, once the sentinel is chosen, the model will generate the token from the GCN-based recommender, which means the model will produce the top-ranked venue name as the generated token. Otherwise, the model chooses a token in $V$ as the decoded token. Basically, the sentinel token is used as a hard gate to control where the next token is generated from at each time step. In this way, we do not need to separately learn a gating function as in \cite{gulcehre2016pointing}. Also, our model is not constrained by a soft gate mechanism as in \cite{see2017get}.

\subsection{Training Objectives}
\label{train}
As the generation of responses is controlled via the sentinel token $\$$ as a hard gate, the generation procedure actually works in a two-step way. The substitution of $\$$ with venue recommendation result is separate from the token generation process. In order to achieve good results, we train the whole model in a sequential way. At the beginning, we train the global topic control component separately on the altered dataset where all venue names are substitute with $\$$. The training objective of this component is $\mathcal{L}_{Topic}$ detailed as Equation \ref{topic}.

Then we change back the dataset and train the GCN component for venue ranking on it. The dialog context is embedded via the trained global topic control model. The training objective is $\lambda \mathcal{L}_{GCN}$ as detailed in Equation \ref{gcn}.

Finally, we initialize the whole model with the components trained and fine-tune them altogether. The final training objective is as follows,
\begin{align*}
\mathcal{L} = \mathcal{L}_{Topic} + \lambda \mathcal{L}_{GCN},
\end{align*}
where $\lambda$ is the weight to balance the losses of the two components. In our experiments, we empirically set this hyperparameter to $0.1$.

\section{Experiments}
In this section, we systematically evaluate the proposed method, termed as \textbf{DCR}, in travel. The experiments are carried out to answer the research questions as follows. 

\begin{itemize}
	\item[RQ1:] Can the proposed DCR properly respond to users' queries in travel? What are the key reasons behind?
	\item[RQ2:] Does the topic control component help the system generate coherent responses? Are the learnt topics reasonable?
	\item[RQ3:] Does the GCN-based recommender help the system find appropriate venues? Whether the relationships between venues are important to capture?
\end{itemize}
In what follows, we will first describe the experimental settings. We then answer the above three research questions.

\subsection{Experimental Setup}
\subsubsection{Dataset}
Arguably the greatest bottleneck for statistical approaches to dialog system development is the collection of appropriate training dataset, and this is especially true for task-oriented dialog systems \cite{liao2018knowledge}. Fortunately, \cite{budzianowski2018multiwoz} contributed a dataset consisting of over 10K conversation sessions in travel domain --- MultiWOZ, which is a fully-labeled collection of human-human written conversations. During the collection of this dataset, it simulates natural conversations between a tourist and a clerk from an information center in a touristic city. Various possible dialog scenarios are considered, ranging from requesting basic information about attractions through booking a hotel room or traveling between cities. In total, the presented corpus consists of 7 sub-topics --- Attraction, Hospital, Police, Hotel, Restaurant, Taxi, Train. The dialogs cover between 1 and 5 sub-topics per dialog thus greatly varying in length and complexity. This broad range of topics captures the common scenarios where sub-tasks are naturally connected in travel. For example, a tourist needs to find a hotel, to get the list of attractions and to book a taxi to travel between both places.

In total, there are 10, 438 dialogs collected, where 3, 406 of them focus in single-topic dialogs and 7,032 of them are dialogs consisting of at least 2 up to 5 sub-topics. In the experiment, we follow random split of train, test and development set in the original paper. The test and development sets contain 1k examples each. Generally, around 70\% of dialogs have more than 10 turns which shows the complexity of the corpus. The average number of turns are 8.93 and 15.39 for single and multi-domain dialogs respectively with 115, 434 turns in total. The average sentence lengths are 11.75 and 15.12 for users and system response respectively. The responses are also more diverse thus enabling the training of more complex generation models.

\subsubsection{Comparing Methods}
To evaluate the effectiveness of the proposed method, we compare it with the following state-of-the-art solutions. 

\begin{itemize}
	\item[--] \textbf{HRED} \cite{serban2016building}: It predicts the system utterance given the history utterances. The history is modeled with two RNNs in two levels: a sequence of tokens for each utterance and a sequence of utterances. This model works as the basis for our method and other baselines.
	\item[--] \textbf{MultiWOZ} \cite{budzianowski2018multiwoz}: It frames the dialog as a context to response mapping problem, a seq2seq model is augmented with an oracle belief tracker and a discrete database accessing component as additional features to inform the word decisions in the decoder. Note that a seq2seq model is used in the original paper, we extend it to HRED to model multi-turn dialogs. 
	\item[--] \textbf{Mem2Seq} \cite{madotto2018mem2seq}: It augments the existing MemNN \cite{sukhbaatar2015end} framework with a sequential generative architecture to produce coherent responses for task-oriented dialog systems. It uses global multi-hop attention mechanisms to copy words directly from dialog history or KBs. 
	\item[--] \textbf{TopicRNN} \cite{dieng2016topicrnn}: It incorporates topic information into the seq2seq framework to generate informative and interesting responses for chatbots. We also extend the encoder part to model multi-turn dialogs.
	\item[--] \textbf{ReDial} \cite{li2018towards}: It integrates the HRED based conversational model with a denoising auto-encoder based recommender \cite{sedhain2015autorec} via a switching mechanism. The recommendation part is pre-trained separately and only considers the co-occurrence of items while ignores the dialog context. The recommender part is also compared in ablation study.
	\item[--] \textbf{NCF} \cite{he2017neural}: It employs deep learning to model the key factor in collaborative filtering --- the interaction between user and item features, and achieves good performance. The inner product is replaced with a neural architecture. We compare this recommender with our GCN-based recommender in the ablation study.
\end{itemize}

\subsubsection{Evaluation Protocols}
We evaluate the methods in various evaluation protocols. Due to the difficulty in evaluating conversational agents \cite{liunot2016}, a human evaluation is usually necessary to assess the performance of the models. Therefore, we perform both corpus-based evaluations and human evaluations.
For corpus-based evaluations, we adopt the BLEU score and Entity Accuracy as our evaluation metrics, where:
\begin{itemize}
	\item[--] \textbf{BLEU}: Being commonly used in machine translation evaluations, BLEU score has also been widely used in evaluating dialogs systems \cite{eric2017copy}. It is based on the idea of modified n-gram precision, where the higher score denotes better performance. 
	\item[--] \textbf{Entity Accuracy}: Similar to \cite{madotto2018mem2seq}, we average over the entire set of system responses and compare the entities in plain text. The entities in each gold system response are selected by a predefined entity list. This metric evaluates the ability to recommend appropriate items from the provided item set and to capture the semantics of dialogs \cite{eric2017copy}.
\end{itemize}

For human evaluations, we define a set of subjective scores to evaluate the performance of various methods. We run a user study to assess the overall quality of the responses of our model as compared to the baselines. To do a less biased evaluation, we recruit five participants and present each of them ten generated dialog sessions from our test set. The participants are asked to give Fluency scores and Informativeness scores for the generated system responses. They are also asked to provide the rankings of each method for each dialog session. We allow ties so that multiple methods could be given the same rank for the same dialog session (e.g., rankings of the form 1, 2, 2, 2, 2 are possible if the one method is clearly the best, but the other four are of equivalent quality).

\begin{itemize}
	\item[--] \textbf{Fluency}: It evaluates how fluent the generated responses are. The score ranges from zero to five, where a larger score indicates the generated response is more fluent.
	\item[--] \textbf{Informativeness}: This score shows whether the generated responses are informative or not, or say whether users' queries get properly answered. It also ranges from zero to five, where a larger score indicates that the evaluator thinks that the generated response is more informative.
	\item[--] \textbf{Ranking}: This metric directly shows how good each method is as compared to the others. It reflects the overall feeling of users regarding the performance of each method.
\end{itemize}

\subsubsection{Training Setups}
The proposed model is implemented in PyTorch \footnote{Our code will be made publicly available for reproducibility.}. We use the provided development set to tune the hyper-parameters, track the training progress and select the best performing model for reporting the results on the test sets. The components of the joint architecture are first trained separately to achieve a relatively good performance. We then combine them together and fine-tune by minimizing the sum of various loss functions as detailed in Section \ref{train}. We use an embedding size of 300, GRU state size of 100. The embeddings are initialized from pretrained GloVe embeddings \cite{pennington2014glove} and fine-tuned during training. We use two layers of graph convolutional operations. Mini-batch SGD with a batch size of 64 and Adam optimizer \cite{Kingma14} with a learning rate of 0.01 is used for training.

We use the Python-based natural language toolkit NLTK to perform tokenization. Entities in dialog sessions are recognized via heuristic rules plus database entries. All counts, time and reference numbers are replaced with the $\left<\text{value\_count}\right>$, $\left<\text{value\_time}\right>$ and $\left<\text{domain\_reference}\right>$ tokens respectively. To reduce data sparsity further, all tokens are transformed to lowercase letters. The stop words are chosen using tf--idf \cite{blei2006correlated}. The number of topics $K$ is set to 20. All tokens that appear less than 5 times in the corpus are replaced with the $\left<\text{UNK}\right>$ token. We follow the \{S,U,S'\} utterance ``triples'' structure as \cite{serban2016building} in our experiments, which means we aim to generate the system utterance S' by observing the former 1 turn of system utterance S and user utterance U.

\subsection{Performance Comparison}
\subsubsection{Corpus-based Evaluation}
\begin{figure}[!htp]
	\centering
	\includegraphics[scale=0.58]{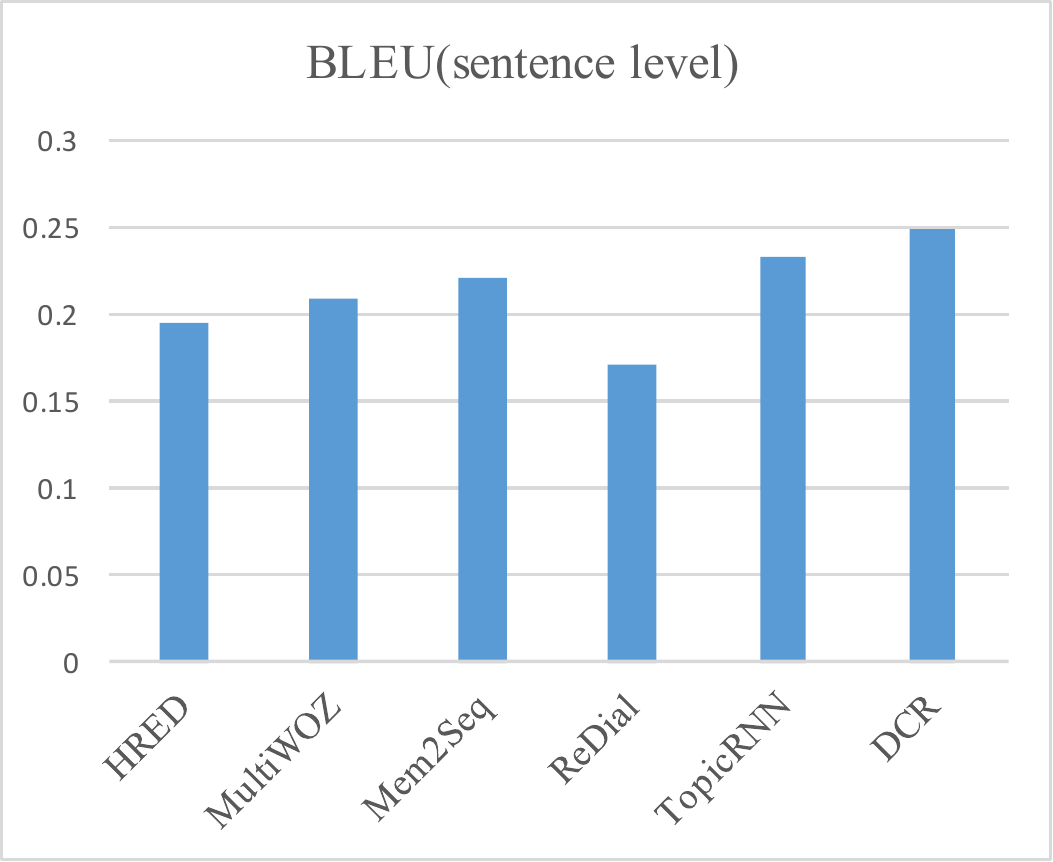}
	\caption{The BLEU scores for each method.}
	\label{Fig:bleu}
\end{figure}

\begin{figure}[!htp]
	\centering
	\includegraphics[scale=0.58]{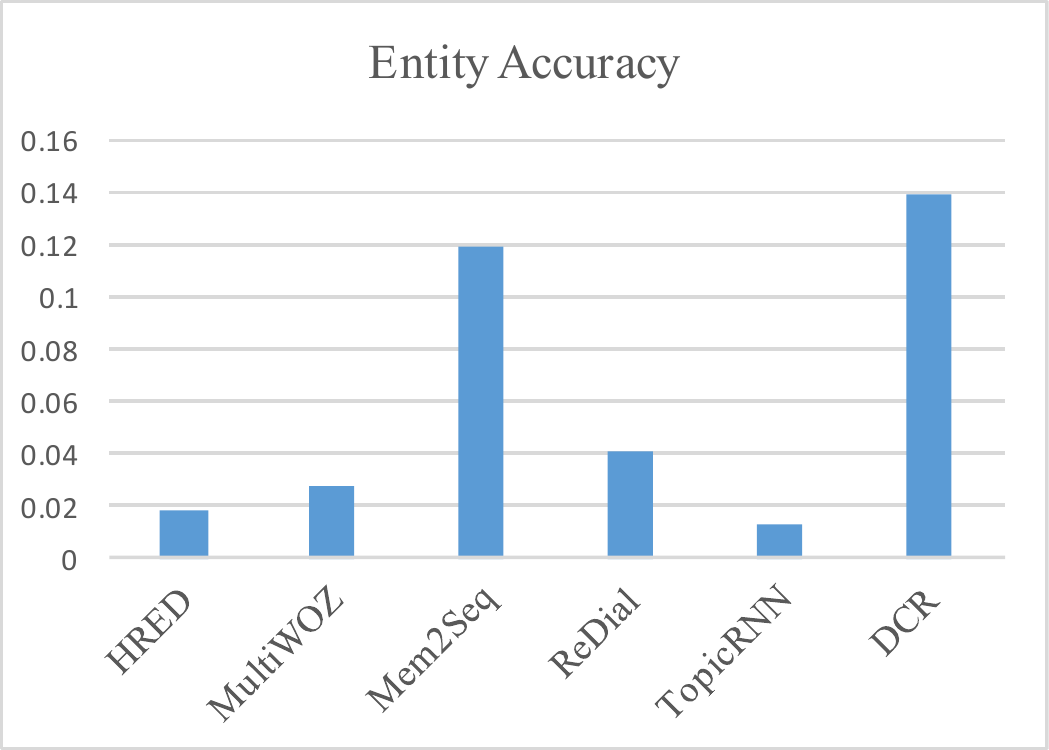}
	\caption{The entity accuracy scores for each method. Note that this is the top-1 accuracy score since only the top ranked venue is leveraged by text response.}
	\label{Fig:acc}
\end{figure}

The result of the corpus-based evaluation is presented in Figure \ref{Fig:bleu} and Figure \ref{Fig:acc}. For each method, the results are obtained based on the best model chosen via the development set. The key observations are as follows.

Overall, the proposed DCR method performs better than all the other baselines in both metrics -- BLEU and entity accuracy. For example, regarding BLEU score, we observe a 6.82\% of performance improvement as compared to the second best method, TopicRNN. The two methods perform better than all the other baselines.
In terms of the entity accuracy score, DCR improves the performance of venue recommendation by 17.2\% as compared to the second best method, Mem2Seq. The performance improvements of DCR method demonstrate its effectiveness in travel domain conversational recommendation due to the following aspects: a) DCR has a global topic control component which enables the system to adaptively generate within topic responses based on the context topic. The learned topics narrow down the generation of tokens in decoding. b) The graph convolution operation incorporates venue information as well as venue relations in the learned venue representations. It matches the venues with the dialog contexts which is essential for conversational recommendation.

In more detail, we analyse the BLEU score shown in Figure \ref{Fig:bleu} first. It reflects the quality of generated text responses. Generally speaking, all methods manage to achieve some improvements over the basic framework -- HRED. For the MultiWoz method, the performance improvement is due to the incorporation of a belief tracker and a discrete database accessing component. However, the improvement is less than that of the Mem2Seq method, because MultiWoz encodes the belief states into anonymous vectors and only the database search count is leveraged. Mem2Seq, on the contrary, generates responses from the dialog history and KB --- some tokens or entities are directly copied to form responses. It happens frequently that words appeared in dialog context are re--used by later responses, which is the underlying reason for its good performance. 
For the method ReDial, since a pointer softmax is leveraged to integrate the text modeling and the recommendation part, its BLEU score might get affected.
When it comes to TopicRNN, we observe a performance improvement, which is mainly attributed to the topic mechanism. It helps to generate tokens matching the dialog context topic and narrow down the generation of tokens. In addition to a similar topic control scheme, DCR manages to achieve superior performance by achieving better entity prediction. 

Regarding the entity accuracy score presented in Figure \ref{Fig:acc}, we observe that the basic end-to-end framework method, HRED, performs rather badly. It is as expected since the method only treat venue entities as tokens and generate tokens based on the encoded dialog context. The basic information of venue entities and the relationships between them are ignored. For the MultiWoz method, although a database query component is leveraged, it only makes use of the number of obtained results. Therefore, the performance is still relatively low. When it comes to the Mem2Seq method, there is a large performance leap. By observing the corpus, we find that the reason might be due to frequent entity re-use phenomenon in dialogs as we detailed before --- venue entities appeared in dialog context will likely to re-appear in the following responses. For the ReDial method, it manages to achieve better performance than that of its basic framework HRED but the improvement is limited. Although it has a denoising autoencoder based recommender, it is largely affected by the data sparsity problem in the dataset, and the recommendations are only conditioned on the entities mentioned in the context but not directly on the language, e.g. texts like ``a cheap restaurant'' in dialog context are ignored. For the TopicRNN method, we also observe a rather low performance on entity prediction. The reason behind is similar to that of the HRED method. On the contrary, the proposed DCR method is able to achieve superior performance on finding the appropriate venue entities. This is because the GCN-based recommender jointly considers the venue information, venue relationships and their match to the dialog context.

\subsubsection{Human Evaluation}
We present the averaged human evaluation results in Table \ref{table:textabl} (the Fleiss' kappa value between evaluators is 0.65). It directly reflects human perception of the quality of generated responses. The results show that the proposed DCR achieves the best performance across these various metrics, which indicates that the responses generated by it are more fluent and informative. We show that the performance improvements of DCR over the other methods are significant. For example, in terms of the Fluency score, DCR improves the performance of response generation by 50.0\%, 44.5\%, 30.3\%, 53.5\% and 8.8\% as compared to the HRED, MultiWoz, Mem2Seq, Redial and TopicRNN methods, respectively. Intuitively, at a certain degree, the BLEU score also reflects how fluent the responses are. In the results, these two metrics indeed show similar pattern of performance improvements. As detailed before, the main reason for the superior performance of DCR might be due to the global topic control mechanism. In travel, dialogs naturally involve multiple sub-tasks, which leads to several topics in the dialog flow. The topic control component enables the system to swiftly switch among topics and generate within-topic responses. 
\begin{table}[!htp]
	\centering
	\footnotesize
	\caption{Human evaluation results for different methods.}
	\label{table:textabl}
	\renewcommand{\arraystretch}{1.2}
	\begin{tabular}{l|c|c|c}
		\hline
		\textbf{Method}&\textbf{Fluency}&\textbf{Informativeness}&\textbf{Ranking}\\[+0.2em]
		\hline
		HRED &2.64&2.34&3.08\\
		\hline
		MultiWOZ &2.74&2.82&2.7\\
		\hline
		Mem2Seq &3.04&3.06&2.3\\
		\hline
		ReDial & 2.58& 2.62&2.8\\
		\hline
		TopicRNN &3.64&2.78&2.66\\
		\hline
		\textbf{DCR} &\textbf{3.96}&\textbf{3.82}&\textbf{1.8}\\
		\hline
	\end{tabular}
\end{table}

At the same time, the Informativeness score shows whether user queries are properly addressed. It not only includes the evaluation of recommended venues but also the information slots appeared in responses such as \textit{food type}, \textit{hotel price} etc. We observe that the general performance pattern resembles that of the entity accuracy metric. However, the Informativeness score of DCR is much larger than that of Mem2Seq. This might be due to the fact that although the venue entities can re-occur in responses, the value of information slots usually require outside knowledge. In DCR, since it already manages to recommend the venue, the slot values are obtained via the venue information through a simple post-process.

For the final ranking of methods, we find it in general accord with the Fluency and Informativeness score trends. The DCR is ranked as the best method by our evaluators, followed by the Mem2Seq method. It actually points out a future direction to enhance our method. Due to the frequent ``re-use'' phenomenon in dialogs, the dialog context is important to model. To encode it into vector representation as in HRED is not sufficient, direct incorporation of the tokens as in the Mem2Seq method opens a new auxiliary road.

\subsection{Analysis on Components}
In this subsection, we explore the performance and contribution of the major components in our design. We first evaluate the global topic control component. We showcase the learned topic words, the stop word indicator prediction results and topic distribution of some example dialogs. Then, we explore the recommendation component. We compare the GCN-based recommender with several state-of-the-art recommendation methods and provide the results.
\begin{table}[!htp]
	\centering
	\footnotesize
	\caption{Four representative topics from the global topic control component.}
	\label{table:topicwords}
	\renewcommand{\arraystretch}{1.3}
	\begin{tabular}{c|c|c|c}
		\hline
		\textbf{Restaurant}&\textbf{Hotel}&\textbf{Attraction}&\textbf{Taxi}\\[+0.2em]
		\hline
		restaurant&hamilton&region&runs\\
		\hline
		eastern&guesthouse&shopping&vehicle\\
		\hline
		cantonese&convenient&modern&departures\\
		\hline
		appeal&stayed&fabulous&campus\\
		\hline
		vegetarian&aylesbray&world&birmingham\\
		\hline
		menu&warkworth&churchhill&arriveby\\
		\hline
		eritrean&accommodation&christ&driving\\
		\hline
		caribbean&arrangements&shopping&causeway\\
		\hline
	\end{tabular}
\end{table}

\subsubsection{Topic Control of Dialogues}
Here we evaluate the performance of the global topic control component. At first, we show whether the learned topic words are coherent. We run the component on our dataset with the total topic number $K$ set to 20. To give a clear view, we show several representative topic words in Table \ref{table:topicwords}. The first row entries indicate the estimated topics for their corresponding column of topic words, where these topic words are top-ranked ones within each column group. Generally speaking, we observe that words are grouped together and the top-ranked words show certain topic meanings within each group.

\begin{figure}[!htp]
	\centering
	\includegraphics[scale=0.4]{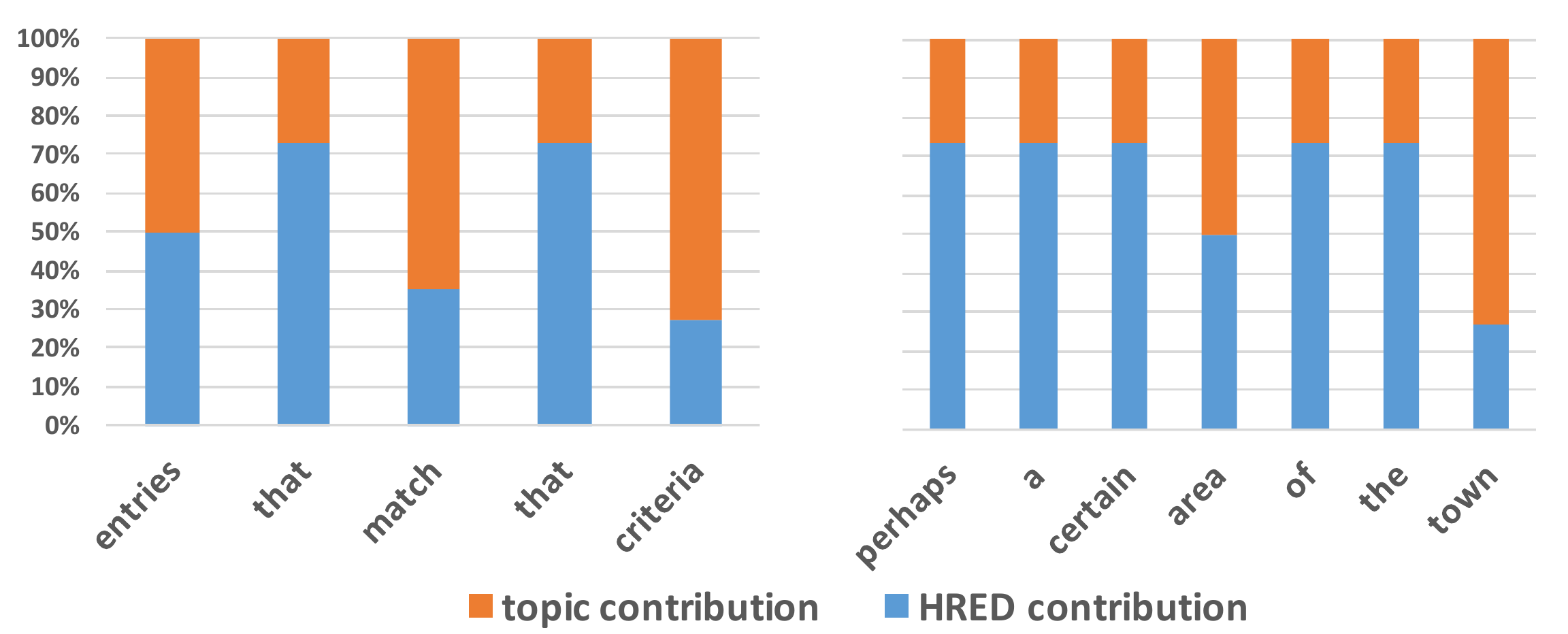}
	\caption{Analysis of the learned stop word indicators.}
	\label{Fig:indicator}
\end{figure}
Since stop words largely affect the performance of our global topic control component, we also show the analysis of the stop word indicator $l$. Some examples are shown in Figure \ref{Fig:indicator}. As can be seen, the learned indicators correspond to the human intuition, and help to coordinate the contribution of the global topic part and the local syntactic part while generating responses.

\begin{figure*}[!htp]
	\centering
	\includegraphics[scale=0.55]{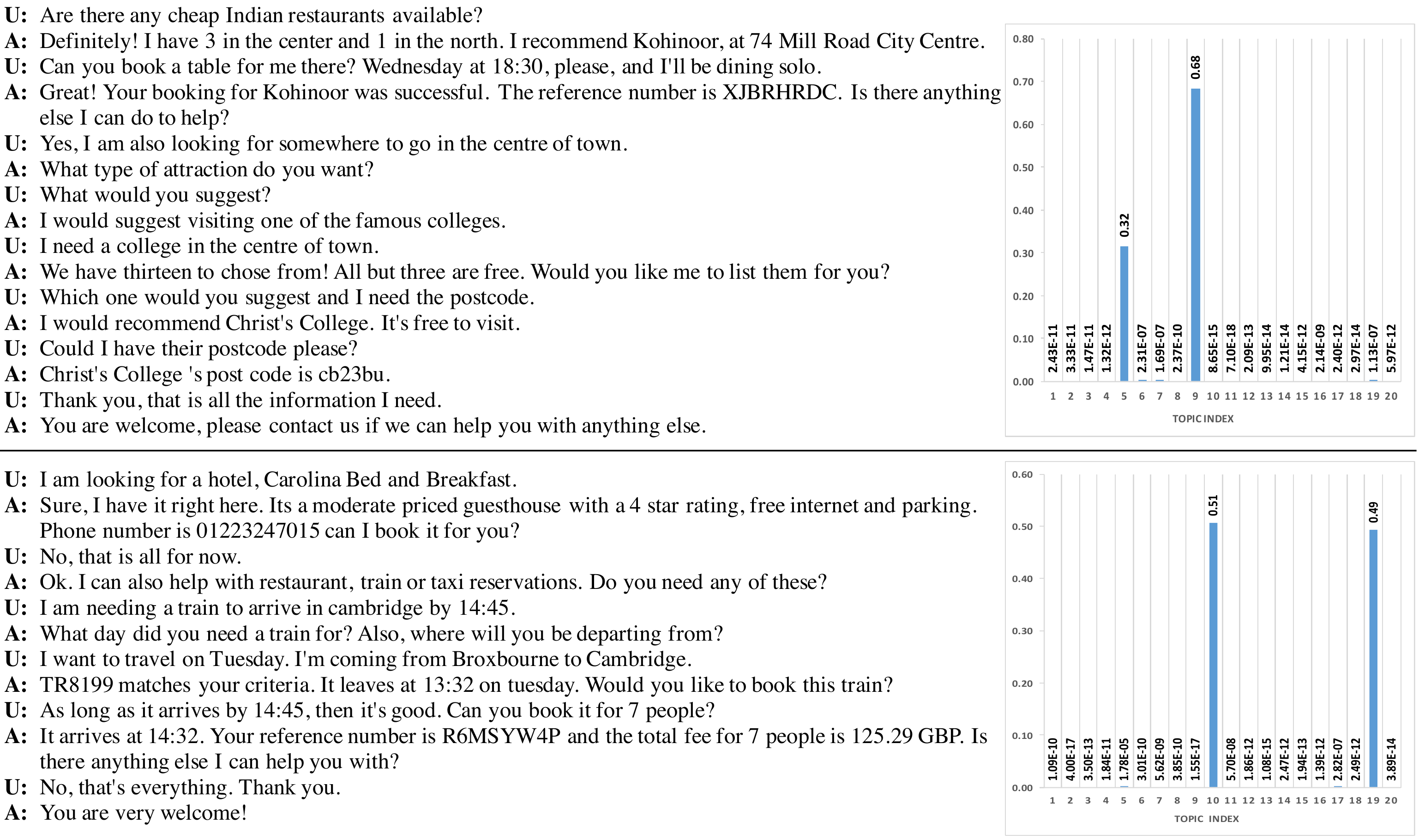}
	\caption{Inferred topic distribution of two example dialog sessions. It shows that some of the topics have been picked up depending on the dialog content.}
	\label{Fig:topicdistri}
\end{figure*}
To show whether the global topic control component correctly captures the overall topic distribution of dialog sessions, we plot the topic distribution $\boldsymbol{\theta}$ of dialogs as shown in Figure \ref{Fig:topicdistri}. Two example dialog session are presented, and both of them are paired with the learned topic distribution. In the first example, the user asks the agent to book a table in a cheap Indian restaurant first, then to recommend an attraction in the center of town. There are two sub-tasks involved during the dialog session. Accordingly, the global topic control component picks up two topics successfully. By observing the top-ranked topic words for each topic indicators, we find that the picked topic 9 refers to \textit{restaurant} while topic 5 refers to \textit{attraction}. It demonstrates the effectiveness of our topic control. Similarly, in the second example, the user first wants to find a hotel and then book the train tickets. There are also two sub-tasks involved in the dialog. Thus, two topics are picked up by the model. We observe that topic 10 corresponds to \textit{hotel} and topic 19 corresponds to \textit{train}.

\subsubsection{Venue Recommendation Analysis}
In this subsection, we analysis our GCN-based venue recommendation component in detail. The common user-item interaction situation is abstracted from the dialogs by treating the dialog contexts as representations of users and venue entities as items. This scenario is used in the NCF method. For ReDial, a user-based autoencoder for collaborative filtering (U-Autorec) is leveraged where venue entities appeared in the same dialog session are extracted to form the entity vector.
\begin{table}[!htp]
	\centering
	\footnotesize
	\caption{Performance comparison of recommenders.}
	\label{table:recommend}
	\renewcommand{\arraystretch}{1.2}
	\begin{tabular}{c|c|c|c}
		\hline
		\textbf{Methods}&\textbf{ReDial}&\textbf{NCF}&\textbf{GCN-based}\\[+0.2em]
		\hline
		Top-1 Accuracy&0.1065&0.1882& \textbf{0.2420}\\
		\hline
	\end{tabular}
\end{table}

The results are shown in Table \ref{table:recommend}. Since often only the top item is leveraged in the dialogs, we report the Top-1 accuracy here. It shows that the GCN-based recommender component achieves better performance as compared to ReDial and NCF methods. For the ReDial recommender, it projects the entity appearance vector $v$ of each dialog session into a smaller vector space, then retrieve a new entity vector $v'$ with same dimension to minimize the difference between them. It only models the co-occurrence relationship among entities. The entity information and the dialog context information are largely ignored. At the same time, the entity co-occurrence matrix formed via training dialog sessions is rather sparse. These factors together lead to its relatively weak performance. Regarding the NCF method, the dialog contexts are gathered via HRED to form vector representations of users. We adopt a multi-layer perceptron (MLP) to learn the interaction between user and item features. Still, the various relationships between venue entities are not modeled. On the contrary, the GCN-based recommender component in DCR manages to handle all the three evidence sources --- the venue information, relations between them and the match to dialog context.

\section{Conclusion}
In order to build an intelligent conversational agent in travel domain, we proposed a deep conversational recommender to answer various user queries. It is equipped with a global topic control component to adaptively generate within-topic responses based on the dialog context topics, which narrows down the generation of tokens in decoding. At the same time, a graph convolutional network based recommender manages to pop venues by modeling the venue information, relations between them and the match to dialog context. Based on the results from the two components, the final response is generated by incorporating them via a pointed integration mechanism. We systematically evaluated the proposed method on a large conversational dataset in travel. Experimental results showed that the proposed DCR method outperformed a wide range of baselines and demonstrated the effectiveness of it in generating fluent and informative responses.

In future, we will continue our work in two directions. First, we will explore the ``re-use'' phenomenon to further boost the performance of response generation. Second, we will try to leverage extra venue adoption data from travel e-commerce sites to enhance the recommendation performance.
\ifCLASSOPTIONcompsoc
  \section*{Acknowledgments}
\else
  \section*{Acknowledgment}
\fi

This research is supported by the NExT++ research center, which is supported by the National Research Foundation, Prime Minister's Office, Singapore under its IRC@SG Funding Initiative. We warmly thank all the anonymous reviewers for their time and efforts.



%




\end{document}